\documentclass[conference]{IEEEtran}
\usepackage{cite}
\usepackage{algorithmic}                
\usepackage{algorithm}                  
\usepackage{amsfonts}                   
\usepackage{amsmath}                    
\usepackage{graphicx}                   
\usepackage{caption}

\ifCLASSINFOpdf
\else
\fi
\begin{document}
%
\title{Multi-modal Feedback for Affordance-driven Interactive Reinforcement Learning}

\author{\IEEEauthorblockN{Francisco Cruz\,$^{1,2}$, German I. Parisi\,$^{1}$ and Stefan Wermter\,$^{1}$}
\IEEEauthorblockA{$^{1}$Knowledge Technology, Department of Informatics, Universit\"at Hamburg, Germany.\\
$^{2}$Escuela de Computaci\'on e Inform\'atica, Facultad de Ingenier\'ia, Universidad Central de Chile, Santiago, Chile.\\
\{cruz,parisi,wermter\}@informatik.uni-hamburg.de}
}


%


\maketitle

\begin{abstract}
Interactive reinforcement learning (IRL) extends traditional reinforcement learning (RL) by allowing an agent to interact with parent-like trainers during a task.
In this paper, we present an IRL approach using dynamic audio-visual input in terms of vocal commands and hand gestures as feedback.
Our architecture integrates multi-modal information to provide robust commands from multiple sensory cues along with a confidence value indicating the trustworthiness of the feedback.
The integration process also considers the case in which the two modalities convey incongruent information.
Additionally, we modulate the influence of sensory-driven feedback in the IRL task using goal-oriented knowledge in terms of contextual affordances.
We implement a neural network architecture to predict the effect of performed actions with different objects to avoid failed-states, i.e., states from which it is not possible to accomplish the task.
In our experimental setup, we explore the interplay of multi-modal feedback and task-specific affordances in a robot cleaning scenario.
We compare the learning performance of the agent under four different conditions: traditional RL, multi-modal IRL, and each of these two setups with the use of contextual affordances.
Our experiments show that the best performance is obtained by using audio-visual feedback with affordance-modulated IRL.
The obtained results demonstrate the importance of multi-modal sensory processing integrated with goal-oriented knowledge in IRL tasks.
\end{abstract}


%
\IEEEpeerreviewmaketitle

\section{Introduction}

Interactive reinforcement learning (IRL) has received increasing attention for teaching autonomous robotic agents to perform a task.
In traditional reinforcement learning (RL), an agent interacts with the environment performing actions and observing new situations to learn an optimal policy which allows it to learn how to perform a task autonomously~\cite{Sutton98}.
Although RL has been shown to be effective for learning agents, one open issue is the excessive time required by the agent to find a proper policy~\cite{Griffith13}.
IRL speeds up this learning process by extending traditional RL with a parent-like teacher who is allowed to advise the agent in selected episodes.

Caregivers interact with infants through different multi-sensory stimuli such as speech and gestures.
These stimuli can also be seen as guidance from a teacher who provides a set of instructions on how to achieve a specific goal.
Although IRL approaches have been implemented in robotic scenarios, an open issue is that the communication interface between the teacher and the robot may not be straightforward for non-expert trainers in a domestic environment.
Therefore, this is motivation to develop simpler interactive scenarios where parent-like teachers can provide instructions using their natural communication skills such as speech and gestures.
In this setting, the feedback provided by the user may, however, be incongruent or noisy.
From a computational perspective, the use of multiple sensory modalities has shown to attenuate the ambiguity of incoming stimuli~\cite{Bauer15}, providing the means to enhance perception-driven behavior.
However, the process of multi-sensory integration must also take into account the case in which the information from the multiple sources is in conflict (incongruent).
Uncertainty in the integration of multi-sensory instructions may not be clear and misunderstood, thereby leading to a decreased performance in the apprentice agent when solving a task~\cite{Griffith13}.

In previous research~\cite{CruzIROS}, we presented a multi-modal IRL approach using dynamic audio-visual input as trainer-like feedback.
Multi-modal feedback can be provided to the agent through a set of vocal commands and hand gestures using a microphone and a depth sensor respectively.
Our IRL algorithm integrates audio-visual information with the aim to provide robust feedback along with a confidence value that indicates the level of trustworthiness of the feedback based on sensory cues.
However, a limitation of this approach is that multi-modal feedback is taken into account on the basis of the confidence value of the integrated feedback predictions, thus relying on the ability of our architecture to robustly predict speech and gestures from sensory cue (which is not always the case in real-world scenarios).

In this paper, we extend our architecture to modulate the influence of sensory-driven feedback in the IRL task using goal-oriented knowledge.
This is motivated by neurobehavioral studies on multi-modal processing in which human subjects exposed to audio-visual stimuli integrate multiple sources of information driven by a combination of sensory representations and prior expectations~\cite{Odegaard15}.
In our approach, we integrate task-specific knowledge in terms of contextual affordances which represent an effective method to anticipate the effect of actions performed by an agent interacting with objects~\cite{CruzESANN}.
We train a multi-layer neural network to predict the effect of the performed actions with different objects in order to avoid failed-states, i.e., states from which it is not possible for the agent to complete the task.
Therefore, multi-modal processing driven by audio-visual sensory representations is integrated with knowledge about the task so that, e.g., predicted feedback with a high confidence value can be bypassed in case that this prediction leads to a failed-state.

We conduct a set of experiments to explore the interplay of external feedback and contextual (task-specific) affordances in a cleaning scenario in which a humanoid robot can interact with two objects with the goal of cleaning the surface of a table.
We compared the learning performance in terms of speed of convergence and accumulated reward under 4 different conditions: traditional RL, multi-modal IRL, and each of these two setups with the use of affordances.
For this purpose, we varied the percentage of available feedback and contextual affordances during the learning process.
The obtained results demonstrate that multi-modal sensory processing integrated with affordance-driven IRL yield the best learning performance for the proposed IRL task.

\begin{figure}
\centering
\includegraphics[width=\linewidth]{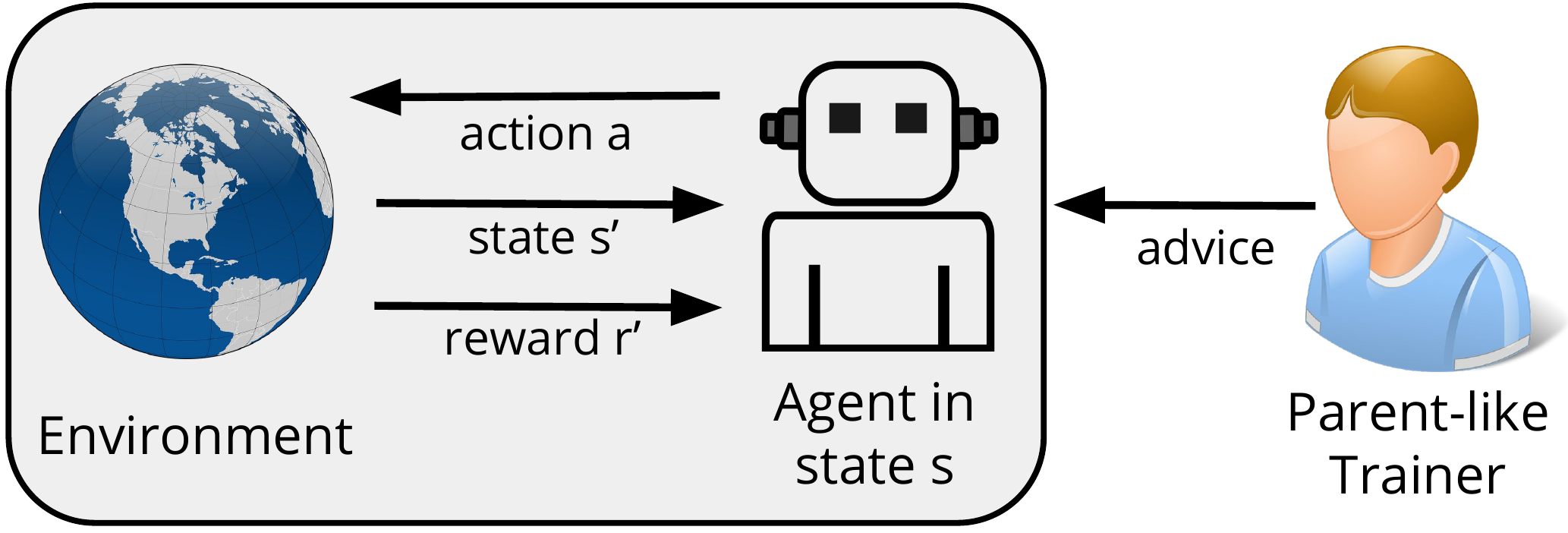}
\caption{Schematic view of the interactive reinforcement learning (IRL) approach.}
\label{fig:IRLScenario}
\end{figure}

\section{Related Work}

\subsection{Interactive Reinforcement Learning}

RL is a behavioral approach used by autonomous agents to learn new tasks~\cite{Sutton98}.
The agent interacts with the environment in order to find a proper policy which determines how to act to accomplish a given task effectively. 
During a learning episode, an RL agent performs an action $a$ obtaining a reward (or a punishment) $r'$ and a new state $s'$ from the environment~(Fig.~\ref{fig:IRLScenario}).
These actions are selected according to a policy $\pi$, which in psychology is referred to as a set of stimulus--response rules or associations~\cite{Kornblum90}.
The value of taking an action $a$ in a state $s$ under a policy $\pi$ is denoted as $q^{\pi}(s,a)$, which is also called the action-value function for a policy $\pi$.

The process of learning in humans and animals has been widely studied by neuroscience, yielding a better understanding of how the brain is able to acquire new cognitive skills. 
RL is associated with cognitive memory and decision-making mechanisms in the biological brain in terms of how behavior is generated~\cite{Niv09}. 
RL is a method used to address optimal decision-making, attempting to maximize the collected reward and minimize the punishment over time, and shown to be successful in terms of acquiring new skills in robotics~\cite{Kober13}\cite{Kormushev13}.

To learn a task, an RL agent has to interact with the environment in order to collect and refine knowledge over time.
Nevertheless, it may be sometimes inefficient to give the agent full autonomy when learning a task since this may lead to excessive training time.
IRL allows to speed up the learning process by using a parent-like advisor (Fig.~\ref{fig:IRLScenario}) to support the learning by delivering useful advice in selected episodes by either reward- or policy-shaping~\cite{Griffith13}.
This approach reduces the search space and allows to learn the task faster in comparison to a fully autonomous agent~\cite{Suay11, Knox13}.
Therefore, an apprentice agent can be taught by a parent-like trainer in a similar way as caregivers assist infants during the learning of new tasks.
A parent-like trainer can be either a human user or another artificial agent. 

In robotics scenarios, IRL has been applied using different communication interfaces between the trainer and the apprentice agent. 
For example, Suay and Chernova~\cite{Suay11} proposed an IRL task where a humanoid robot receives advice from an external trainer using a graphic user interface built on top of the visual input from the robot's camera. 
The trainer may provide feedback through the use of a tablet to deliver directions to the robot.
In another IRL approach by Knox~et~al.~\cite{Knox13}, the trainer can send feedback to the apprentice robot using a presenter control.
However, the communication interfaces lack usability when it comes to home scenarios where non-expert trainers may train assistive robots in a more natural way. 


\subsection{Multi-modal Integration}

In our daily life, we are constantly subject to external stimuli through different sensory modalities (e.g., vision, audition, and touch) that together provide a coherent and robust perceptual experience~\cite{Stein09}.
Similarly, robot perception may be driven by an array of sensors that in concert contribute to the efficient and robust interaction with the environment.
In human-robot interaction (HRI) scenarios, robots may take advantage of multi-sensory information to reliably operate in highly dynamic environments and in situations of sensory uncertainty.

From a computational perspective, multiple studies have shown the advantages of integrating multi-sensory information. 
For instance, Lacheze et al.~\cite{Lacheze09} presented a multi-modal integration approach for the classification of static patterns using audio-visual input. 
This work uses the auditory information to improve the classification of objects when they were partially obstructed.
Kimura et al.~\cite{Kimura15} proposed a multi-modal approach to estimate the characteristics of unknown objects using an RGB-D camera, a stereo microphone, and sensors of pressure and weight. 
Ozasa et al.~\cite{Ozasa12} presented an approach to recognize unknown objects using multi-modal integration of images and speech through logistic regression. 
They also integrated a confidence value to improve the accuracy of the recognition.
However, this integrated confidence value did not take into account the case of conflict of uni-modal information.
Such a conflict may occur due to either sensor noise or incongruent advice provided by the user.
In previous research~\cite{CruzIROS}, we proposed an integration function that considers incongruent audio-visual input so that in the case of conflict, the predicted label with the highest confidence is preferred.
Both of these approaches yield exclusively sensory-driven integration, thus do not consider task-specific knowledge which plays an important role in multi-modal integration~\cite{Odegaard15}.

\subsection{Affordances}

Affordances are available possible actions for an agent operating in its environment~\cite{Gibson79}. 
They represent characteristics of the relation between the agent and an object in terms of operational opportunities the object offers to the agent.
The original idea of affordances by Gibson included many practical examples but lacked a formal definition leading to significant differences among cognitive psychologists about the definition and use of affordances~\cite{Horton12}, and also in the field of artificial intelligence~\cite{Chemero07}.
Horton et al.~\cite{Horton12} recognized three main attributes of an affordance: i) its existence is associated with the capabilities of an agent, ii) it exists regardless whether the agent is able to perceive it, and iii) it does not change, unlike necessities or goals of the agent.
Different approaches for learning affordances in robotics have been proposed.
For instance, Lopes et al. \cite{Lopes07} addressed the imitation learning problem using affordance-based action sequences. 
The affordance model was extended allowing the robot to work with a second object using an enlarged Bayesian network to represent affordances~\cite{Moldovan12}, studying the use of functional affordances which also lead to a reduced action space.
In this regard, an object can be used in a restricted manner by not considering all its operational opportunities but only the socially acceptable ones. 

In robotics, affordances have been represented as a triplet $\mathit{affordance:= <object, action, effect>}$ which encodes relationships between its components~\cite{Montesano08}. 
Therefore, it is possible to predict the effect using objects and actions as domain variables as $\mathit{effect=f(object, action)}$.
Nevertheless, although this model has been shown to be suitable for many scenarios, it does not include contextual information which allows anticipating effects in all situations properly~\cite{Kammer11}. 
For instance, let us consider a scenario where an agent interacts with a set of given objects.
In the case that the agent has his/her two hands occupied with objects, then the agent cannot grasp a new object.
In other words, the affordance of graspability is temporarily unavailable until the agent drops one object or places it back onto the table (which in turn may modify the context).
This aspect must be considered for agents operating in real-world scenarios for an efficient and effective interaction with the environment.

\begin{figure*}[!t]
  \centering
    \includegraphics[width=0.8\textwidth]{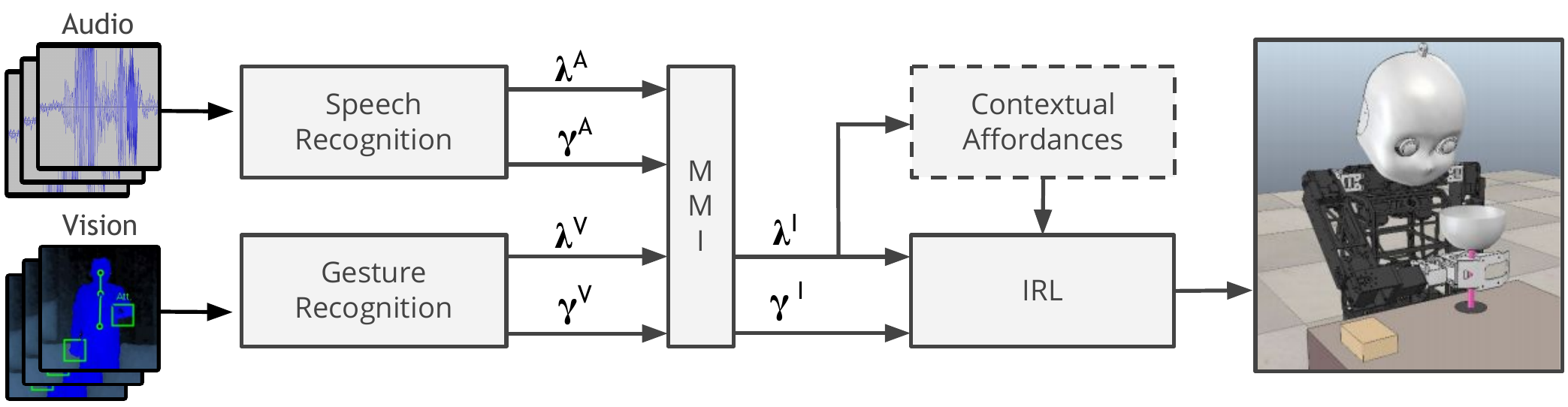}
  \caption{A diagram of our architecture. The multi-modal integration (MMI) module estimates a label $\lambda^I$ and a confidence value $\gamma^I$ from the speech and gesture recognition modules. Integrated feedback is used to compute contextual affordances (the effects of action $\lambda^I$ given the current state). Contextual affordances modulate the influence of external feedback on the IRL algorithm.}
  \label{fig:diagram}
\end{figure*}

\section{Our Method}

A diagram of our architecture is illustrated in Fig.~\ref{fig:diagram}.
Audio-visual input is processed by two distinct modules that recognize feedback commands from speech and gestures.
Both recognition modules predict a command label and a confidence value that indicates the level of trustworthiness of such predictions.
The multi-modal integration (MMI) module computes a joint label $\lambda^I$ and a confidence value $\gamma^I$ on the basis of uni-modal predictions.
For this purpose, we use a mathematical transformation that takes into account also incongruent information, i.e., predicted command labels from speech and vision that do not match.
The labels of the integrated feedback are used as the input to compute the contextual affordances, i.e., the effects of an action $\lambda^I$ given the current state.
Thus, the IRL algorithm may consider or not the feedback, e.g., bypassing the feedback that leads the agent to a failed-state from which the task cannot be successfully carried out.
In the following sections, we describe in detail the above-mentioned modules and their implementation.

\subsection{Speech and Gesture Recognition}


To process speech from auditory information, we used a cloud-based automatic speech recognition (ASR) system with local audio signals.
The ASR system is based on DOCKS~\cite{Twiefel14} that selects the best hypothesis given a domain-specific sentence list associated with our robot task. 
Our domain-specific language is comprised of robot commands represented by a list of sentences which can be interpreted as advice for the agent.
To determine which sentence fits better, we use the Levenshtein distance that compares each obtained hypothesis to the sentences in our domain-specific language using a phonemic representation.
Given the set $H$ of the 10-best hypotheses and the set $S$ of the in-domain sentences, the predicted class label is computed as:
\begin{equation}
\lambda^A~=~\mathrm{argmin}~\mathcal{L}(h_i, s_j),
\end{equation}
where $\mathcal{L}$ is the Levenshtein distance in our ASR system.
The confidence value is computed as: 
\begin{equation}
\gamma^A = \max (0, 1 - \mathcal{L}(h_i, s_j)/|s_j|),
\end{equation}
with $h_i \in H$ and $s_j \in S$, both represented as phonemes.


For recognizing gestures, we used an extended version of neural network learning for gesture recognition \cite{ParisiHandSOM} that extracts hand-independent gesture features from depth map sequences.
The learning model consists of a set of two hierarchically arranged self-organizing networks that learn the spatiotemporal structure of the input sequences in terms of gesture features.
Along with a predicted label, we also estimate a confidence value that expresses the degree of belief that the prediction is correct based on sensory-driven observations.
Training videos were recorded with an ASUS Xtion depth sensor from which we estimated the 3D skeleton model.

A label prediction is carried out every 3 frames in a sliding window scheme.
We consider the last 5 observations and compute the statistical mode that returns the most frequent value given the set of predictions $\Lambda^V$, from which we compute the gesture class as $\lambda^V=Mo(\Lambda^V)$.
Let $N$ be the number of occurrences of $\lambda^V$ in $\Lambda^V$ so that the confidence value can be defined as $\gamma^V=N/|\Lambda^V|$, thus yielding a confidence value in the range between $1$ and $0.2$.

\subsection{Audio-visual Integration}

To integrate the multi-sensory input, we propose a non-linear transformation function.
This function defines the relationship between the predicted labels and the confidence values: $(\lambda^A,\gamma^A)$ for audio and $(\lambda^V,\gamma^V)$ for vision.
We compute the integrated label $\lambda^I$ using the highest confidence value:
\begin{equation}
\lambda^I = \left\{
\begin{array}{c c c}
  \lambda^A& & \mathrm{if~} \gamma^A > \gamma^V\\
  \lambda^V& & \mathrm{otherwise} \\
\end{array}
\right.
\end{equation}


Therefore, when $\lambda^A$ and $\lambda^V$ are equal, any of them is assigned to $\lambda^I$. In case that $\lambda^A$ and $\lambda^V$ are different, then the label with a greater confidence is assigned to $\lambda^I$.
The integrated confidence value is computed by the function $\gamma^I=\ln{(1 + \phi)}$, with $\phi$ being a dynamic parameter that depends on each congruent or incongruent pair of predicted labels and their confidence values.
We refer to this parameter as the \textit{likeliness parameter} which we compute as follows:

\begin{equation}
\phi = \left\{
\begin{array}{c c c}
  \gamma^A + \gamma^V & & \mathrm{if~} \lambda^A = \lambda^V\\
  |\gamma^A - \gamma^V| &  & \mathrm{if~} \lambda^A \neq \lambda^V\\
\end{array}
\right.
\label{eq:eta}
\end{equation}

The likeliness parameter strengthens the integrated confidence value $\gamma^I$ when the predicted labels for audio and vision are congruent, whereas $\gamma^I$ diminishes if the predicted uni-modal labels are incongruent.
After applying the transformation function, the confidence value is rescaled between $0$ and $1$. 


\subsection{Affordance-driven IRL}

To introduce task-specific information to the available affordances, we use contextual affordances where an additional variable is considered denoting the current state~\cite{CruzESANN}.
In this case, the affordance triplet $\mathit{affordance:= <object, action, effect>}$ is now extended to $\mathit{contextualAffordance:= <state, object, action, effect>}$, yielding:
\begin{equation}
\mathit{effect=f(state, object, action)},
\end{equation}
where $\mathit{state}$ is the agent's current state, $\mathit{object}$ is one of the entities that the agent can interact with, $\mathit{action}$ represents motor behavior that can be triggered by the interaction with the objects, and the $\mathit{effect}$ is the outcome of a specific action involving agent-object interaction \cite{Atil10}.

The model of contextual affordances that includes the agent's current state enables to provide knowledge about actions that lead to failed-states from which it is not possible to accomplish the given task and, therefore, the action space is reduced by avoiding these states.

Formally, the problem can be stated as follows.
Given an agent performing the same action $a$ with the same object $o$ but from a different agent's state $s_1 \neq s_2$: when the action $a$ is performed, different effects $e_1 \neq e_2$ could be generated since the initial states $s_1$ and $s_2$ are different. 
It is then unfeasible to establish differences in the final effect when we use affordances to represent it because $e_1=(a,o)$ and $e_2=(a,o)$. 
Thus, to deal with the current states $s_1 \neq s_2$, an agent must learn to distinguish each case by using contextual affordances defined by $e_1=(s_1,a,o)$ and $e_2=(s_2,a,o)$, which establishes clear differences between the final effects.


The described model of contextual affordances allows us to predict the effects of specific actions, in our case with the use of a feed-forward neural network that learns the relationship of the states, the actions, and the objects.
In our IRL approach, we use different levels of availability for contextual affordances to modulate the learning process.
Contextual affordances are used in both autonomous RL and IRL, i.e., the selected action (either by the agent autonomously or by the parent-like trainer as feedback) may be bypassed if the effect of performing such an action leads the agent to a failed-state.


\section{Experimental Results}

\subsection{Robot Scenario}
\label{RoSce}

In order to test our method, we implement a robotic home scenario where a robot interacts with a parent-like trainer to perform a cleaning task.
The task consists of wiping a table with the use of one of the robot's arms.
To successfully complete the cleaning task, it is necessary to carry out additional sub-tasks such as interacting with objects on the table.
The trainer is able to advise the robot on what action to perform next through the use of speech, gestures, or a combination of both.
The scenario comprises three different locations:
\begin{enumerate}
	\item[i.] \textit{left}, the left section of the table;
	\item[ii.] \textit{right}, the right section of the table;
	\item[iii.] \textit{home}, an additional position that is the initial and final position of the robot's arm.
\end{enumerate}

There are two objects included in the scenario:
\begin{enumerate}
	\item[i.] \textit{sponge}, used to wipe both sections of the table. 
The sponge is placed at the home position while it is not being used by the robot;
	\item[ii.] \textit{goblet}, initially placed in one of the sections of the table and, therefore, it must be moved from one section to the other during cleaning in order to end the task successfully.
\end{enumerate}

The robot can perform seven different actions. 
These actions may be decided by the robot autonomously as well as by the parent-like trainer through multi-modal feedback in terms of audio-visual advice. 

Available actions and advice classes are as follows:
\begin{enumerate}
	\item[i.] \textit{go left}: moves the arm to the left section of the table;
	\item[ii.] \textit{go right}: moves the arm to the right section of the table;
	\item[iii.] \textit{go home}: moves the arm to the home location;
	\item[iv.] \textit{grasp}: takes the object at the current arm's position;
	\item[v.] \textit{place}: drops the object at the current arm's position;
	\item[vi.] \textit{wipe}: uses the sponge to wipe the table at the current arm's position;
	\item[vii.] \textit{abort}: aborts the task and returns to the initial state.
\end{enumerate}

Each state is represented by four variables:
\begin{enumerate}
	\item[i.] \textit{handObject}: the object which is currently in the robot's hand (sponge, goblet, or free); 
	\item[ii.] \textit{handPosition}: the position of the robot's arm (left, right, or home); 
	\item[iii.] \textit{gobletPosition}: the position of the goblet on the table (left or right); 
	\item[iv.] \textit{sideCondition}: a vector with two values indicating whether all sections of the table have been wiped. 
\end{enumerate}
Therefore, the state vector can be represented as follows:
\begin{equation}
s_t=<handObj,handPos,gobletPos,sideCondition>.
\end{equation}

To complete the task, the robot must clean both sections of the table by moving the goblet from one section to the other during the process of wiping. 
After the robot has wiped both sections, the task finishes with the sponge at the home position and with free hands.
Therefore, the final state $s_f$ can be represented as:
\begin{equation}
s_f= <free, home, left | right, [clean, clean]>.
\end{equation}

%

Once the task has been finished and the final state has been reached, the agent receives a reward of $1$. 
If the agent is not able to finish the task due to a failed-state, it receives a negative reward (or punishment) of $-1$. 
In all other states, the agent receives a small negative reward of $-0.01$ for each transition in order to discourage longer paths and loops. 
The reward function is summarized as follows:
\begin{equation}
r(s) = \left\{
\begin{array}{r l}
  1 & \textrm{if the agent finishes}\\
 -1 & \textrm{if the agent cannot continue}\\
 -0.01 & \textrm{otherwise.}
\end{array}
\right.
\label{eq:reward}
\end{equation}

RL is performed using SARSA with a discount factor $\gamma=0.9$, learning rate $\alpha=0.3$, and $\epsilon$-greedy action selection with $\epsilon=0.1$.
IRL is carried out using a probability of feedback of $0.3$, meaning $30\%$ of the time we use advice to assist the robot in the task execution.

Our architecture comprises four different modules.
The interface module is in charge of receiving the parent-like advice using a depth sensor and a microphone. 
The control module receives the uni-sensory elements of advice to compute multi-modal feedback, which is sent to the learning algorithm and to the affordance module to predict possible failed-states. 
Finally, the robot module generates low-level action control using either a real or a simulated robot. 

Our approach uses contextual affordances to predict the effect after an action has been performed by the robot in the cleaning-table scenario.
We encode all the variables with a localist data representation for objects, locations, side conditions, and actions. 
We use this representation to create the training set for the contextual affordance model which is composed of a multi-layer perceptron. 
As input, we use vectors with $20$ components containing information about the current state and the action.
The current state is represented by the first 13 components in the input vector considering the four variables that define a state, i.e., hand object, hand position, goblet position, and side condition. 
The output corresponds to the effect of contextual affordances encoded as a vector with 13 components representing the next state.
If the performed action leads to a failed-state, then all components of the output vector are equal to zero.
The training data was created considering all possible combinations of states together with actions, i.e., $53$ states and $7$ actions.
Therefore, the total number of data samples is $371$ for the training of the multi-layer perceptron with $30$ hidden neurons with a sigmoid transfer function using the Levenberg-Marquardt learning algorithm~\cite{Hagan94} for $100$ epochs.


\subsection{Results and Evaluation}

\begin{figure}
\begin{center}
\includegraphics[width=0.85\linewidth]{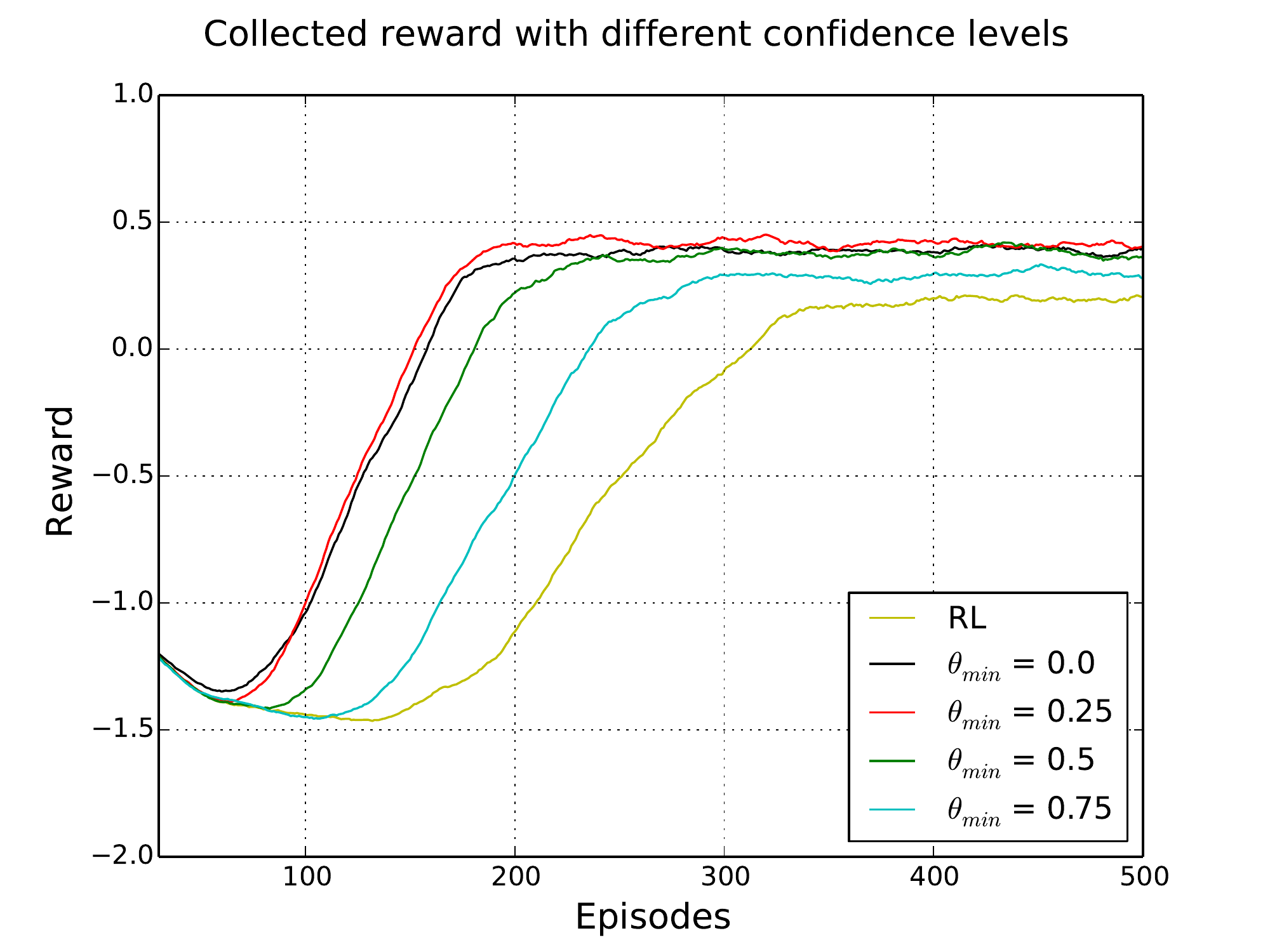}
\end{center}
\caption{Collected rewards using autonomous RL and IRL with multi-modal feedback. 
The best performance was obtained for $\theta_{min}=0.25$ (red line).}
\label{fig:rewardsIntegrated}
\end{figure}

We implemented the cleaning table scenario described in Sec.~\ref{RoSce} in order to test our proposed method.
For this purpose, we made recordings of speech and hand gesture sequences from a parent-like teacher.
These recordings enabled us to better control the conducted experiments in order to repeat the process under different learning conditions.
Each advice class was recorded four times.
After the training was completed, our goal was to predict the feedback labels from novel multi-modal input sequences $(\lambda^A,\lambda^V)$ along with their confidence values $(\gamma^A,\gamma^V)$.
After processing each modality independently, the predictions were integrated using the multi-modal integration model to compute $\lambda^I$ and $\gamma^I$.
We considered $\gamma^I>\theta_{min}$ with $\theta_{min}$ as the minimum confidence value to be considered as a valid advice. 
Then, we used different $\theta_{min}$ to verify whether smaller confidence values are still beneficial. 
The thresholds used were $\theta_{min} \in \{0.0, 0.25, 0.5, 0.75\}$. 
The average convoluted rewards are shown in Fig. \ref{fig:rewardsIntegrated} for 100 agents and 500 episodes.

\begin{figure}
\centering
\includegraphics[width=0.85\linewidth]{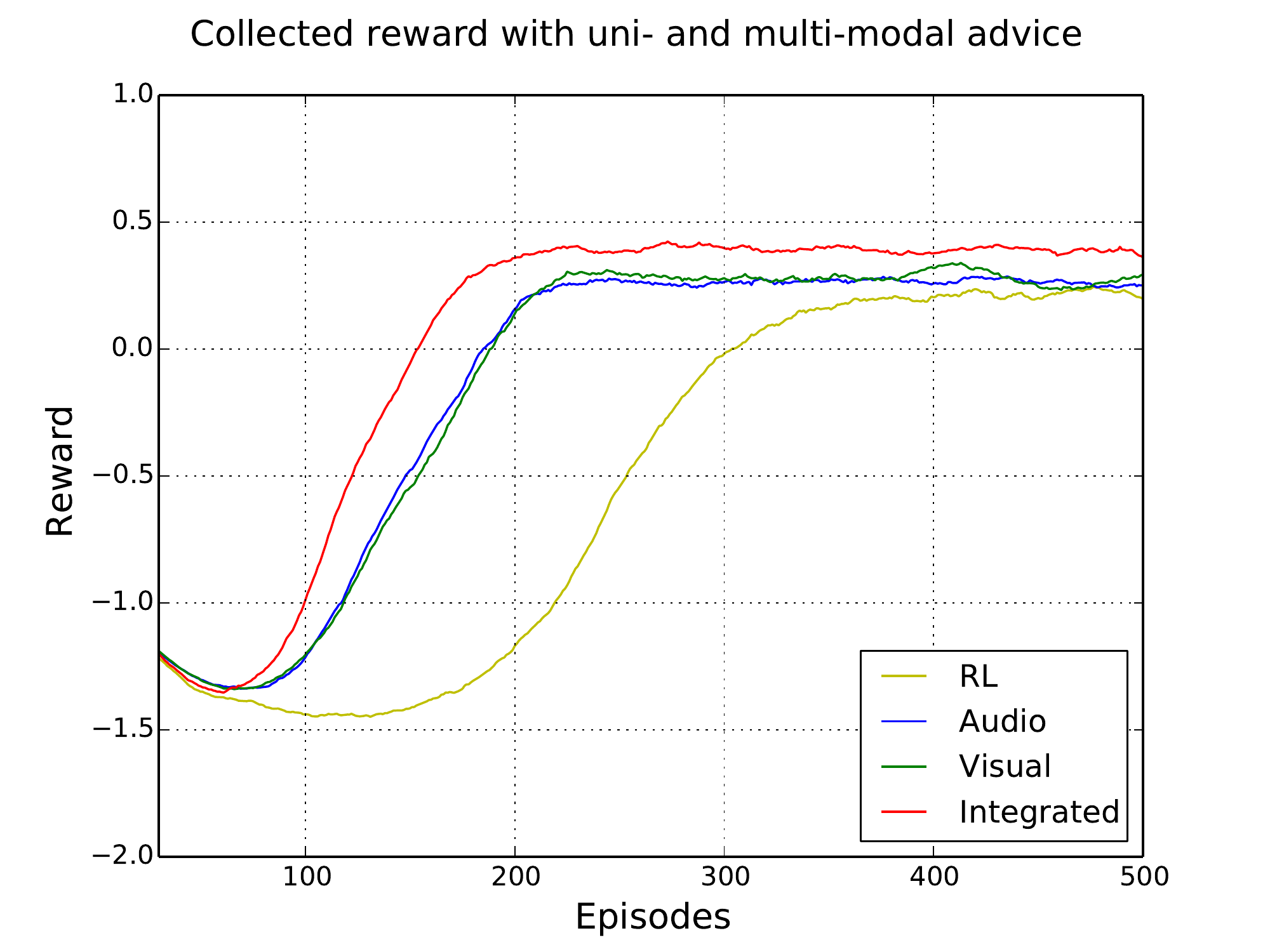}
\caption{Collected rewards using autonomous RL and IRL with uni-modal feedback, and IRL with multi-modal feedback ($\theta_{min}=0.25$).
The multi-modal IRL approach converges faster and to a greater reward than uni-modal approaches.}
\label{fig:rewardsCompared}
\end{figure}

To evaluate differences in the uni-modal and multi-modal approaches, we used a threshold of $\theta_{min}=0.25$. 
The results for $100$ agents and $500$ episodes are shown in Fig.~\ref{fig:rewardsCompared} where it is possible to observe that both uni-modal approaches lead to similar learning behavior, i.e., similar convergence speed and accumulated reward. 
When using integrated multi-modal advice, the approach converges faster and collects greater reward in comparison with audio and visual advice only.

\begin{figure*}
\centering
\includegraphics[width=0.81\linewidth]{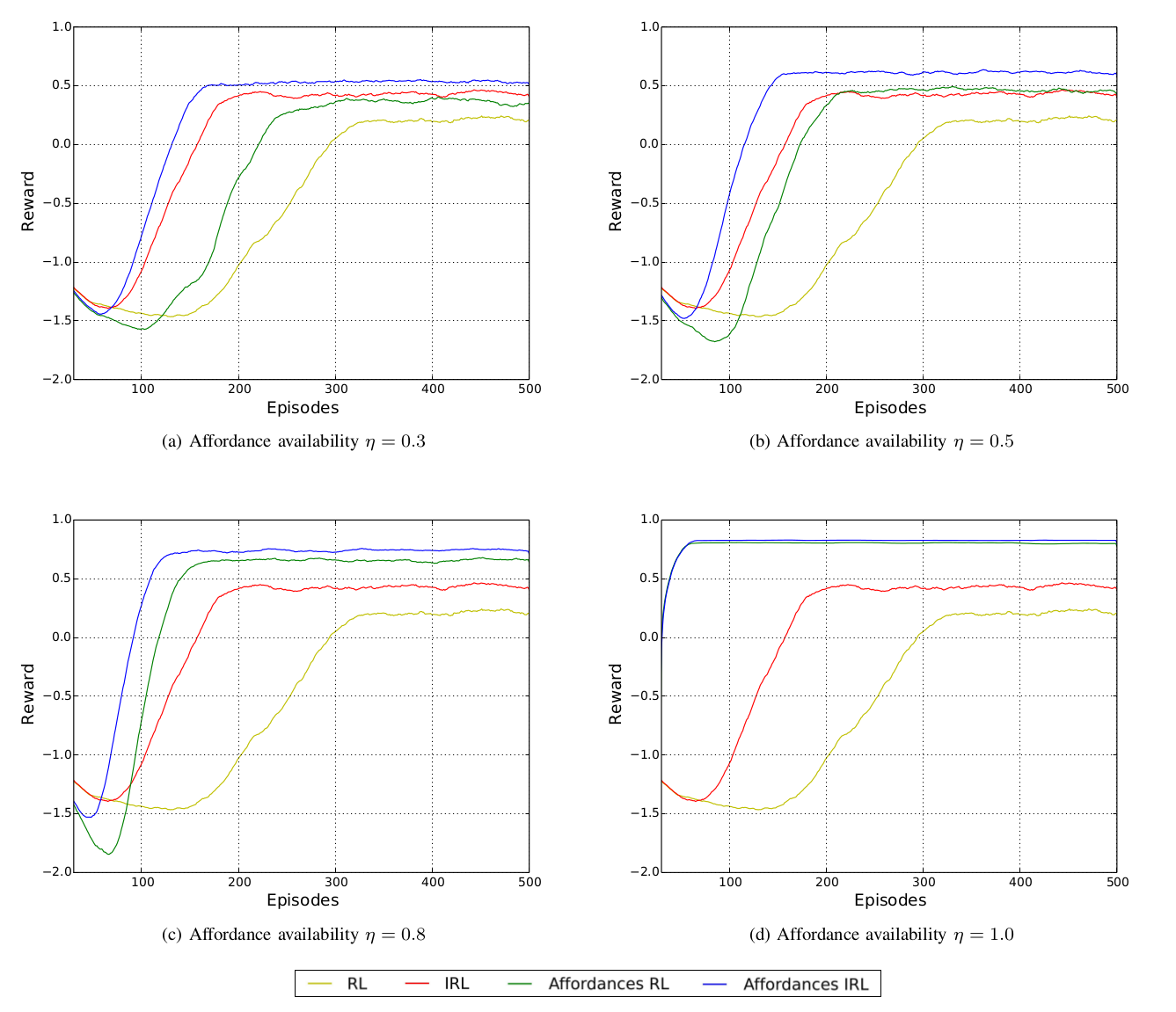}
\caption{The plots show the collected reward for different values of affordance availability using autonomous RL and IRL. The results are compared with respect to RL and IRL without using contextual affordances. In all cases, the affordance-driven approaches yield a better performance in terms of the collected reward and convergence speed.}
\label{fig:rewardsAffordance}
\end{figure*}

Afterwards, we introduced the proposed contextual affordance model to avoid failed-states. 
This way, we do not only reduce the action space but also the likelihood of a failed-state during an episode so that the agent is less likely to receive a punishment. 
Consequently, using the affordance-driven IRL increases the average accumulated reward and yields faster convergence (shown in Fig.~\ref{fig:rewardsAffordance}).
For a better comparison of our results, all plots in Fig.~\ref{fig:rewardsAffordance} show the autonomous RL and IRL approaches without the use of affordances.

We evaluated our method using different percentages of available contextual affordances during the learning process. 
We defined a parameter $\eta$ to be the likelihood of a contextual affordance being available. 
We set values for $\eta \in \{0.3, 0.5, 0.8, 1.0\}$ with $\eta=1$ meaning that the affordance is fully available.
It can be seen in Fig.~\ref{fig:rewardsAffordance}a that even a reduced amount of affordance availability ($\eta=0.3$) improves the learning process.
Furthermore, the autonomous affordance-driven RL approach (Fig.~\ref{fig:rewardsAffordance}, green line) accomplishes similar performance to IRL without affordances in terms of accumulated reward. 
In the case of affordance-driven IRL, it reaches a better performance than IRL without affordances.
Fig.~\ref{fig:rewardsAffordance}b shows the results for an affordance availability of $\eta = 0.5$.
In this case, even the affordance-driven autonomous RL approach obtains a higher accumulated reward in comparison to the IRL approach where affordances are not used. 
With $\eta=0.8$, both approaches with affordances outperform the traditional RL and IRL approaches in terms of accumulated reward and convergence speed (Fig.~\ref{fig:rewardsAffordance}c).
Finally, we used an agent with full affordance availability, i.e., $\eta=1.0$. 
Fig.~\ref{fig:rewardsAffordance}d shows that with affordances being fully available, the agent quickly converges to its maximal possible reward in both RL and IRL approaches, with a slight difference in the maximal reward for both approaches.

\section{Discussion}

\subsection{Summary}

We presented an IRL approach using dynamic audio-visual input as feedback in terms of vocal commands and hand gestures for a robotic cleaning task.
Our approach integrates uni-sensory cues to provide multi-modal feedback.
The multi-modal integration module computes a joint label $\lambda^I$ and confidence value $\gamma^I$ on the basis of uni-modal predictions.
The integration process is of particular importance when the two modalities convey incongruent information, i.e., feedback classes predicted by the modules of speech and gesture recognition do not match.
Therefore, our integration function takes into account the confidence level of the predictions to provide the IRL algorithm with consistent feedback.
As an extension of previous research on multi-modal IRL~\cite{CruzIROS}, we implemented a model using contextual affordances for modulating the influence of sensory-driven feedback in the IRL task through goal-oriented knowledge.
We predict the effects of an action $\lambda^I$ given the current state to avoid failed-states.
In this way, the IRL algorithm may consider or not the feedback by bypassing the feedback that leads the agent to a failed-state, thus speeding up the learning process in terms of required episodes for achieving convergence.

Our results in a simulated robot environment show that although uni-sensory modalities show satisfactory prediction accuracy, the use of multi-modal feedback leads to a better performance in our domestic cleaning table scenario in terms of the accumulated reward and required learning episodes.
We evaluated the learning performance under four different conditions: traditional RL, multi-modal IRL, and these two setups with the use of affordances, showing that the best performance is obtained using multi-modal feedback with affordance-driven IRL.


\subsection{Sensory-driven Feedback vs Goal-oriented Knowledge}

The focus of our study was the interplay of multi-modal feedback with task-specific knowledge.
Our previous results showed that integrated audio-visual representations yield more robust feedback for an IRL task with respect to uni-modal approaches~\cite{CruzIROS}.
In particular, audio-visual integration provides the means to solve conflicts, i.e., situations in which predicted feedback labels from the auditory and the visual modules are incongruent.
This supports the idea of multi-modal integration as a method to enhance robot perception and interaction~\cite{Kimura15}.

In our approach, the integration is carried out taking into account the predicted labels and the confidence values from uni-modal cues.
In the case of incongruent audio-visual predictions, the modality yielding the higher confidence value will be preferred.
Gesture labels are predicted by the neural network processing of hand motion features, whereas vocal commands are predicted using in-domain automatic speech recognition.
Consequently, these two approaches provide robust feedback predictions with confidence values computed as a function of a fully sensory-driven process, i.e., a high confidence value indicates that it is very likely that the feedback perceived by the agent matches the one actually given by the trainer.
This procedure, however, does not give any information on whether the piece of feedback is correct or not in terms of the next actions required to accomplish the task.

Signals from multiple sources are combined in the brain taking into account a combination of the reliability of low-level sensor representations and the expectations of an agent in a specific situation (e.g., in terms of task-oriented knowledge)~\cite{Odegaard15}.
Therefore, we integrated this aspect to our previous model in order to study the combination of sensory-driven multi-modal feedback and goal-oriented knowledge in the context of our IRL task.
In our new extended proposed architecture, we integrated task-specific knowledge in terms of contextual affordances which represent an effective method to anticipate the effect of actions performed by an agent interacting with objects based on its current state~\cite{Cruz16, Cruz14}.
We trained a neural network to predict the effect of performed actions with different objects in order to avoid states from which it is not possible for the agent to complete the cleaning task.
Thus, contextual affordances modulate the influence of multi-modal feedback in the IRL algorithm, i.e., if an action provided by the trainer leads to a failed-state, it may be bypassed irrespective of a high (sensory-driven) confidence value.

\subsection{Future Work}

The obtained results motivate the extension of our approach in several directions.
So far, the integration function considers two cues for predicting multi-modal feedback and computing its confidence.
On the one hand, we could think of naturally extending our function to consider input from additional sensory sources, e.g., RGB information as an additional visual cue.
It has been shown that combining depth and RGB information leads to a better recognition accuracy with respect to using a single cue~\cite{cad60_3}.
In the case of our robotic task, conflicting input in terms of incongruent predictions from the auditory and visual modules may be solved by considering multiple visual cues.
On the other hand, we could think of extending our approach with additional modalities, e.g., haptics.
In such a setting, parent-like feedback may be delivered to the agent by providing haptic feedback to its actuators, e.g. moving its arm to grasp an object.


\subsection{Conclusion}

Multi-modal IRL allows the agent to interact in a more natural way with parent-like trainers for dynamically acquiring and refining task-specific knowledge with respect to traditional IRL approaches.
Together, our results demonstrate the contribution of multi-modal sensory processing integrated with goal-oriented knowledge to significantly enhance the interaction between users and agents in robotic learning tasks.

\subsection*{Acknowledgements}

\thanks{\small{The authors gratefully acknowledge partial support by Comisi\'on Nacional de Investigaci\'on Cient\'ifica y Tecnol\'ogica (CONICYT) scholarship 5043 and the German Research Foundation DFG under project CML (TRR 169).}



%
%
%

\bibliographystyle{ieeetr}
\bibliography{biblio}

\begin{thebibliography}{10}

\bibitem{Sutton98}
R.~S. Sutton and A.~G. Barto, {\em Reinforcement Learning: An Introduction}.
\newblock Cambridge, MA, USA: Bradford Book, 1998.

\bibitem{Griffith13}
S.~Griffith, K.~Subramanian, J.~Scholz, C.~Isbell, and A.~Thomaz, ``Policy
  shaping: Integrating human feedback with reinforcement learning,'' in {\em
  Proceedings of Advances in Neural Information Processing Systems},
  pp.~2625--2633, 2013.

\bibitem{Bauer15}
J.~Bauer, J.~D\'avila-Chac\'on, and S.~Wermter, ``Modeling development of
  natural multi-sensory integration using neural self-organisation and
  probabilistic population codes,'' {\em Connection Science}, vol.~27,
  pp.~358--376, 2015.

\bibitem{CruzIROS}
F.~Cruz, G.~Parisi, J.~Twiefel, and S.~Wermter, ``Multi-modal integration of
  dynamic audiovisual patterns for an interactive reinforcement learning
  scenario,'' in {\em Proceedings of the IEEE/RSJ International Conference on
  Intelligent Robots and Systems (IROS)}, pp.~759--766, 2016.

\bibitem{Odegaard15}
B.~Odegaard, D.~Wozny, and L.~Shams, ``Biases in visual, auditory, and
  audiovisual perception of space.,'' {\em PLoS Computational Biology},
  vol.~11, no.~12, 2015.

\bibitem{CruzESANN}
F.~Cruz, G.~I. Parisi, and S.~Wermter, ``Learning contextual affordances with
  an associative neural architecture,'' in {\em Proceedings of the 24th
  European Symposium on Artificial Neural Networks, Computational Intelligence
  and Machine Learning (ESANN)}, pp.~665--670, 2016.

\bibitem{Kornblum90}
S.~Kornblum, T.~Hasbroucq, and A.~Osman, ``Dimensional overlap: Cognitive basis
  for stimulus-response compatibility -- a model and taxonomy,'' {\em
  Psychological Review}, vol.~97, pp.~253--270, 1990.

\bibitem{Niv09}
Y.~Niv, ``Reinforcement learning in the brain,'' {\em Journal of Mathematical
  Psychology}, vol.~53, pp.~139--154, 2009.

\bibitem{Kober13}
J.~Kober, J.~A. Bagnell, and J.~Peters, ``Reinforcement learning in robotics: A
  survey,'' {\em The International Journal of Robotics Research}, vol.~32,
  pp.~1--37, 2013.

\bibitem{Kormushev13}
P.~Kormushev, S.~Calinon, and D.~Caldwell, ``Reinforcement learning in
  robotics: Applications and real-world challenges,'' {\em Robotics}, vol.~2,
  pp.~122--148, 2013.

\bibitem{Suay11}
H.~B. Suay and S.~Chernova, ``Effect of human guidance and state space size on
  interactive reinforcement learning,'' in {\em Proceedings of IEEE
  International Symposium on Robot and Human Interactive Communication RO-MAN},
  pp.~1--6, 2011.

\bibitem{Knox13}
W.~Knox, P.~Stone, and C.~Breazeal, ``Training a robot via human feedback: A
  case study,'' in {\em Proceedings of the International Conference on Social
  Robotics (ICSR)}, pp.~460--470, 2013.

\bibitem{Stein09}
B.~E. Stein, T.~R. Stanford, and B.~A. Rowland, ``The neural basis of
  multisensory integration in the midbrain: Its organization and maturation,''
  {\em Hearing Research}, vol.~258, no.~1–2, pp.~4--15, 2009.

\bibitem{Lacheze09}
L.~Lacheze, Y.~Guo, R.~Benosman, B.~Gas, and C.~Couverture, ``Audio/video
  fusion for objects recognition,'' in {\em Proceedings of IEEE/RSJ
  International Conference on Intelligent Robots and Systems IROS},
  pp.~652--657, 2009.

\bibitem{Kimura15}
D.~Kimura and O.~Hasegawa, ``Estimating multimodal attributes for unknown
  objects,'' in {\em Proceedings of International Joint Conference on Neural
  Networks IJCNN}, pp.~1--8, 2015.

\bibitem{Ozasa12}
Y.~Ozasa, Y.~Ariki, M.~Nakano, and N.~I. Martinetz, ``Disambiguation in unknown
  object detection by integrating image and speech recognition confidences,''
  in {\em Proceedings of Asian Conference on Computer Vision}, pp.~85--96,
  2012.

\bibitem{Gibson79}
J.~J. Gibson, {\em The Ecological Approach to the Visual Perception of
  Pictures}.
\newblock Boston, MA, USA: Houghton Mifflin, 1979.

\bibitem{Horton12}
T.~E. Horton, A.~Chakraborty, and R.~S. Amant, ``Affordances for robots: A
  brief survey,'' {\em AVANT: Journal of Philosophical-Interdisciplinary
  Vanguard}, vol.~3, pp.~70--84, 2012.

\bibitem{Chemero07}
A.~Chemero and M.~T. Turvey, ``Gibsonian affordances for roboticists,'' {\em
  Adaptive Behavior}, vol.~15, pp.~473--480, 2007.

\bibitem{Lopes07}
M.~Lopes, F.~S. Melo, and L.~Montesano, ``Affordance-based imitation learning
  in robots,'' in {\em Proceedings of IEEE/RSJ International Conference on
  Intelligent Robots and Systems IROS}, pp.~1015--1021, 2007.

\bibitem{Moldovan12}
B.~Moldovan, P.~Moreno, M.~van Otterlo, J.~Santos-Victor, and L.~D. Raedt,
  ``Learning relational affordance models for robots in multi-object
  manipulation tasks,'' in {\em Proceedings of IEEE International Conference on
  Robotics and Automation ICRA}, pp.~4373--4378, St. Paul, MN, USA, 2012.

\bibitem{Montesano08}
L.~Montesano, M.~Lopes, A.~Bernardino, and J.~Santos-Victor, ``Learning object
  affordances: From sensory-motor coordination to imitation,'' {\em IEEE
  Transactions on Robotics}, vol.~24, pp.~15--26, 2008.

\bibitem{Kammer11}
M.~Kammer, T.~Schack, M.~Tscherepanow, and Y.~Nagai, ``From affordances to
  situated affordances in robotics -- why context is important,'' in {\em
  Frontiers in Computational Neuroscience (Conference Abstract: Joint IEEE
  International Conference on Developmental Learning and Epigenetic Robotics
  ICDL-EpiRob)}, 2011.

\bibitem{Twiefel14}
J.~Twiefel, T.~Baumann, S.~Heinrich, and S.~Wermter, ``Improving
  domain-independent cloud-based speech recognition with domain-dependent
  phonetic post-processing,'' in {\em Proceedings of Association for the
  Advancement of Artificial Intelligence Conference AAAI}, pp.~1529--1535,
  2014.

\bibitem{ParisiHandSOM}
G.~I. Parisi, D.~Jirak, and S.~Wermter, ``{HandSOM - N}eural clustering of hand
  motion for gesture recognition in real time,'' in {\em Proceedings of the
  IEEE International Symposium on Robot and Human Interactive Communication
  (RO-MAN), Edinburgh, Scotland, UK}, pp.~981--986, 2014.

\bibitem{Atil10}
I.~At{\i}l, N.~Da\u{g}, S.~Kalkan, and E.~\c{S}ahin, ``Affordances and
  emergence of concepts,'' in {\em Proceedings of 10th International Conference
  on Epigenetic Robotics: Modeling Cognitive Development in Robotic Systems},
  pp.~149--156, 2010.

\bibitem{Lapeyre14}
M.~Lapeyre, P.~Rouanet, J.~Grizou, S.~N'Guyen, A.~L. Falher, F.~Depraetre, and
  P.-Y. Oudeyer, ``Poppy: Open source 3d printed robot for experiments in
  developmental robotics,'' in {\em Proceedings of Joint IEEE International
  Conferences on Development and Learning and Epigenetic Robotics ICDL-EpiRob},
  pp.~173--174, 2014.

\bibitem{Hagan94}
M.~T. Hagan and M.~B. Menhaj, ``Training feedforward networks with the
  marquardt algorithm,'' {\em IEEE Transactions on Neural Networks}, vol.~5,
  pp.~989--993, 1994.

\bibitem{Cruz16}
F.~Cruz, S.~Magg, C.~Weber, and S.~Wermter, ``Training agents with interactive
  reinforcement learning and contextual affordances,'' {\em IEEE Transactions
  on Cognitive and Developmental Systems}, vol.~8, pp.~271--284, 2016.

\bibitem{cad60_3}
B.~Ni, Y.~Pei, P.~Moulin, and S.~Yan, ``{Multilevel depth and image fusion for
  human activity detection},'' {\em IEEE Transactions on Cybernetics}, vol.~43,
  no.~5, pp.~1383--1394, 2013.

\end{thebibliography}

\end{document}